\documentclass[conference]{IEEEtran}

\IEEEoverridecommandlockouts
\usepackage{cite}
\usepackage{amsmath,amssymb,amsfonts}
\usepackage{algorithmic}
\usepackage{graphicx}
\usepackage{textcomp}
\def\BibTeX{{\rm B\kern-.05em{\sc i\kern-.025em b}\kern-.08em
    T\kern-.1667em\lower.7ex\hbox{E}\kern-.125emX}}

\usepackage[dvipsnames]{xcolor}
\usepackage[hyphens]{url}
\usepackage[normalem]{ulem}
\usepackage{mathptmx}
\usepackage{algorithm}
\usepackage{xspace}
\usepackage{mathtools}
\usepackage{pifont}
\usepackage{color}
\usepackage{soul}
\soulregister\cite7
\soulregister\ref7
\soulregister\fig7
\soulregister\todo7
\soulregister\sout7
\soulregister\pageref7
\usepackage{color}
\usepackage{colortbl}
\usepackage{comment}
\usepackage{multirow}
\usepackage{multicol}
\usepackage{arydshln}
\usepackage{makecell}
\usepackage{wrapfig,lipsum}
\usepackage{tikz}
\usepackage{subfloat}
\usepackage[caption=false]{subfig}
\usepackage[font=small,labelfont=bf]{caption}
\usepackage{hyperref}
\newcommand{\ignore}[1]{}
\newcommand{\todo}[1]{\textcolor{red}{#1}\xspace}

\newcommand{\old}[1]{}
\newcommand{\fig}[1]{Figure~\ref{#1}}
\newcommand{\sect}[1]{Section~\ref{#1}}
\newcommand{\tab}[1]{Table~\ref{#1}}

\newcommand{\eqn}[1]{Equation~\ref{#1}}
\newcommand{\proposed}[0]{Pre-gated MoE\xspace}
\newcommand{\gpuonly}[0]{GPU-only\xspace}
\newcommand{\gpucpu}[0]{CPU-GPU\xspace}

\newcommand{\ondemand}[0]{MoE-OnDemand\xspace}
\newcommand{\prefetch}[0]{MoE-Prefetch\xspace}

\newcommand{\x}[0]{$\times$\xspace}

\makeatletter
\renewcommand\p@subfigure{\thefigure}
\makeatother
\usetikzlibrary{fit,calc}

\usepackage{listings} 
\usepackage{xcolor}
\definecolor{lightgray}{gray}{0.9}
\begin{document}

\title{\huge Pre-gated MoE: An Algorithm-System Co-Design for\\
Fast and Scalable Mixture-of-Expert Inference\\

\thanks{
\\
\IEEEauthorrefmark{1} Co-first authors who contributed equally to this research.\\
\IEEEauthorrefmark{2} Work done during an internship at Microsoft Research.\\
\noindent\rule{4cm}{0.4pt}

This is the author preprint version of the work. The authoritative version will appear in the Proceedings of the 51st IEEE/ACM International Symposium on Computer Architecture (ISCA-51), 2024.
}
}

\makeatletter
\newcommand{\linebreakand}{%
  \end{@IEEEauthorhalign}
  \hfill\mbox{}\par
  \mbox{}\hfill\begin{@IEEEauthorhalign}
}
\makeatother

\author{
    \IEEEauthorblockN{Ranggi Hwang\IEEEauthorrefmark{1}\IEEEauthorrefmark{2}}
    \IEEEauthorblockA{KAIST\\
    ranggi.hwang@kaist.ac.kr}
\and
    \IEEEauthorblockN{Jianyu Wei\IEEEauthorrefmark{1}\IEEEauthorrefmark{2}}
    \IEEEauthorblockA{USTC / Microsoft Research\\
    noob@mail.ustc.edu.cn}
\and
    \IEEEauthorblockN{Shijie Cao}
    \IEEEauthorblockA{Microsoft Research\\
    shijiecao@microsoft.com}
\\
\linebreakand 
    \IEEEauthorblockN{Changho Hwang}
    \IEEEauthorblockA{Microsoft Research\\
    changhohwang@microsoft.com}
\and
    \IEEEauthorblockN{Xiaohu Tang\IEEEauthorrefmark{2}}
    \IEEEauthorblockA{Microsoft Research\\
    v-xiaohutang@microsoft.com}
\and
    \IEEEauthorblockN{Ting Cao}
    \IEEEauthorblockA{Microsoft Research\\
    ting.cao@microsoft.com}
\and
    \IEEEauthorblockN{Mao Yang}
    \IEEEauthorblockA{{Microsoft Research}\\
    maoyang@microsoft.com}
}

\maketitle

\begin{abstract}

Large language models (LLMs) based on transformers have made significant strides in recent years, the success of which is driven by scaling up their model size. 
Despite their high algorithmic performance, the computational and memory requirements of LLMs present unprecedented challenges. To tackle the high compute requirements of LLMs, the Mixture-of-Experts (MoE) architecture was introduced which is able to scale its model size without proportionally scaling up its computational requirements. Unfortunately, MoE's high memory demands and dynamic activation of sparse experts restrict its applicability to real-world problems.  Previous solutions that offload MoE's memory-hungry expert parameters to CPU memory fall short because the latency to migrate activated experts from CPU to GPU  incurs high performance overhead. Our proposed Pre-gated MoE system effectively tackles the compute and memory challenges of conventional MoE architectures using our algorithm-system co-design. Pre-gated MoE employs our novel pre-gating function which alleviates the dynamic nature of sparse expert activation, allowing our proposed system to address the large memory footprint of MoEs while also achieving high performance. We demonstrate that Pre-gated MoE is able to improve performance, reduce GPU memory consumption, while also maintaining the same level of model quality. These features allow our Pre-gated MoE system to cost-effectively deploy large-scale LLMs using just a single GPU with high performance. 

\end{abstract}

\begin{IEEEkeywords}
Mixture-of-expert, inference system, machine learning, large language model, memory offloading
\end{IEEEkeywords}
\section {Introduction}
\label{sect:intro}

Machine learning (ML) applications based on large language models (LLMs) have taken the world by storm, widely being deployed in various consumer facing products~\cite{chatgpt,codegen,txt2img}.
The success of LLMs has been driven by scaling up the model capacity (i.e., the model size) and its training dataset, the largest trained model size increasing by around 1,000$\times$ within the past 5 years, from a few hundred million parameters to approaching a trillion parameter scale~\cite{gpt3, palm, megatron_lm}.
With larger model size bringing higher model accuracy, it is likely that future models will also increase in their model capacity. However, a critical challenge in sustainably growing model size is its increasingly demanding computation requirement. 

To tackle the high compute requirements of LLMs, 
the Mixture-of-Experts (MoE)~\cite{moe} model  was suggested as an alternative to the previous \emph{dense} LLMs~\cite{t5, gpt3, palm, megatron_lm}. The power of MoE comes from its ability to scale up the model capacity by increasing the number of \emph{expert} parameters within an MoE block. Despite the increase in model parameter size, however, MoE utilizes a \emph{gate function} to 
only partially activate the experts in a \emph{sparse} manner, allowing them to achieve sub-linear compute cost with respect to model capacity. In contrast, prior dense LLMs activate the \emph{entire} model parameters for inference and cause its compute cost to scale quadratically to model size, incurring significant computation overhead.
Despite its merits, a {\bf critical challenge of MoE} is its large memory requirement and the dynamically activated sparse experts which cause high deployment cost, rendering MoE's applicability in real-world problems to be limited.

\begin{enumerate}
\item {\bf Large memory requirement of experts.} While sparsely activating model parameters (i.e., the experts) helps reduce compute cost to achieve the same model quality as their dense LLM counterparts, MoEs require significantly larger memory to accommodate the large number of experts.
For instance, an MoE-based Google SwitchTransformer~\cite{switchtransformer} can have up to $75\times$ more parameters than the FLOPs-equivalent dense T5 model~\cite{t5}.
In other words, MoEs have a much lower \emph{memory-efficiency} vs. dense LLMs, bringing critical system-level challenges in deploying MoE-based LLMs.
To accommodate MoE's large model size under GPU's limited memory size, multiple GPUs can be utilized for deploying MoE where \emph{expert parallelism} is employed to distribute the expert parameters across the GPUs to store only a portion of the experts for inference~\cite{gshard, dsmoe, semoe_training}.

\item {\bf Dynamic and sparse activation of experts.}
Although multi-GPU solutions can distribute expert parameters across the GPU memory, MoE only partially activates a subset of the experts in a sparse manner~\cite{moe}.
This makes the number of experts actually utilized per each GPU to become either very small or non-existent (i.e., none of the experts in a GPU are activated, leaving GPU idle)~\cite{linamoe}.
Furthermore, the sparse expert activation is dynamically decided at runtime, making it difficult to anticipate how many experts will be activated in each GPU.
As such, the effective computation conducted over each GPU becomes low, exhibiting low GPU compute utilization and aggravating the total cost of ownership (TCO) for deployment.

\end{enumerate}

{\bf Prior work} on deploying MoE seeks to address these dual challenges by \emph{offloading} MoE's memory hungry expert parameters into CPU memory or SSD~\cite{dsinf, semoe, semoe_training, meta_moe} (referred to as \emph{MoE-offload} below). The benefit of MoE-offload is that it reduces the number of GPUs required for deploying MoE, which helps increase the GPU's compute efficiency for inference. Offloading MoE parameters, however, is no silver bullet as it comes with a significant increase in inference latency, deteriorating quality of service (QoS) to end users. 
This is because CPU offloading can only resolve MoE's large memory requirement without addressing the \emph{data dependency} issue that arises with dynamic sparse activation, a unique characteristic of the MoE.
In an MoE block, there exists a sequential dependency between  (1) the ``selection'' of which experts should be activated (using MoE's gate function), and (2) the ``execution'' of activated experts for inference. Because such data dependency is \emph{dynamically} resolved by the input data, the two-stage process of (1) \emph{expert selection} and (2) \emph{expert execution} must be serialized back-to-back. Consequently, the latency to migrate the activated expert parameters from CPU to GPU cannot be hidden under MoE-offload and causes severe performance overheads, failing to address the aforementioned challenges of deploying MoE (\sect{sect:offloading}).

In this work, we propose \emph{\proposed}, an algorithm-system co-design that enables MoE inference to incur low GPU memory consumption while still achieving high performance, substantially reducing  TCO. We briefly summarize the {\bf key contribution and novelty} of our \proposed below.

\begin{itemize}
\item {\bf (Algorithm)} In conventional MoE architectures, the gate function in the $N$-th MoE block selects the experts to activate which will then be executed within the \emph{same} $N$-th MoE block. In our proposed design, we modify the role of a gate function to \emph{preemptively} select the experts to be activated for the \emph{next} MoE block (hence its new name, the \emph{pre-gate} function).  More concretely, the pre-gate function in the $N$-th MoE block selects the experts to activate for the ($N$+1)-th MoE block. The novelty of our pre-gate function lies in its ability to completely \emph{eliminate} the sequential dependency between the expert selection and expert execution stage within any given MoE block (i.e., data dependency now exists across the $N$-th MoE block's expert selection and the ($N$+1)-th block's expert execution), which our proposed system effectively utilizes for performance optimization as detailed below.

\item{\bf (System)} Similar to prior MoE-offload systems, our \proposed stores the memory capacity limited expert parameters in CPU memory and reduces the number of GPUs required for inference. Unlike MoE-offload, our \proposed utilizes the pre-gate function to overlap the CPU$\rightarrow$GPU expert migration latency with the expert execution stage, minimizing the expert migration's impact on performance. Specifically, \proposed utilizes the $N$-th pre-gate function to identify the set of experts to activate for the ($N$+1)-th MoE block, in advance, effectively \emph{prefetching} only the activated experts to the GPU in preparation for the ($N$+1)-th block's execution while concurrently going through the expert execution for the $N$-th MoE block. 

\end{itemize}

We evaluate our \proposed system using state-of-the-art MoE models, achieving comparable or even higher model accuracy compared to the original MoE model across a wide range of natural language processing (NLP) tasks (e.g., summarization, and question answering).
At the same time, decoupling the expert selection vs. expert execution stage provides our \proposed to significantly reduce end-to-end inference latency, only adding $23\%$ performance overhead than the oracular, performance-optimal \gpuonly solution that can store the entire MoE parameters in GPU memory. \proposed also reduces peak GPU memory consumption by $4.2\times$ vs. \gpuonly, allowing the deployment of larger LLMs within a single GPU.
Overall, our \proposed system presents a fast, scalable, and cost-effective solution for serving MoE-based LLMs.

\section{Background}
\label{sect:background}

\subsection{Dense LLMs using Transformers}
\label{sect:dense_llm}

\textbf{Transformer model architecture.}
Transformer models~\cite{transformer} have become the dominant approach in designing ML applications for natural language processing (NLP), due to their ability to capture long-range dependencies and complex patterns in data~\cite{bert, megatron_lm}.
There are two primary ways in which a transformer model is structured: an encoder-decoder architecture~\cite{t5} and a decoder-only architecture~\cite{gpt3}.
An encoder-decoder architecture consists of an encoder module that processes the input data (sequence of input tokens) and a decoder module that generates the output (an output token per decoder). The decoder-only architecture on the other hand implicitly incorporates the encoding process within the transformer blocks, eliminating the need for a separate encoder module. Regardless of which architecture is employed, both encoder-decoder and decoder-only architectures employ the following key components of transformers: the self-attention layer, the position-wise feed-forward networks (FFN) layer, normalizations, and residual connections, as shown in~\fig{fig:dense}.
Self-attention helps determine the inter-word relationships and their dependencies within a sequence, whereas the FFN layer applies non-linear transformations to capture complex patterns in the input data.
Both self-attention and FFN account for a significant portion of computation as well as memory requirements of transformer models.

\textbf{Challenges in scaling dense LLMs.}
The success of transformer based dense LLMs has primarily been driven by scaling up the model's capacity (i.e., model size) by stacking a series of transformer blocks~\cite{llm_scaling_0, llm_scaling_1}, providing higher model accuracy.
However, a key challenge in sustainably growing model capacity is its increasingly demanding compute and memory cost, for both training and inference. In particular, as the size of the LLM increases, the demands on compute and memory grow \emph{quadratically}, making it challenging to fit the model within the memory constraints of modern GPUs while also maintaining high compute efficiency~\cite{mt_lng}.
Furthermore, the energy costs associated with training these LLMs are increasing significantly, raising serious concerns on the environmental impact of training and serving LLMs~\cite{sustainable_ai, carbon_emission_nn}.

\begin{figure}[t!]%
\label{fig:transformer}%
\centering
\subfloat[Dense transformer block.]{%
\label{fig:dense}%
\includegraphics[width=0.480\textwidth]{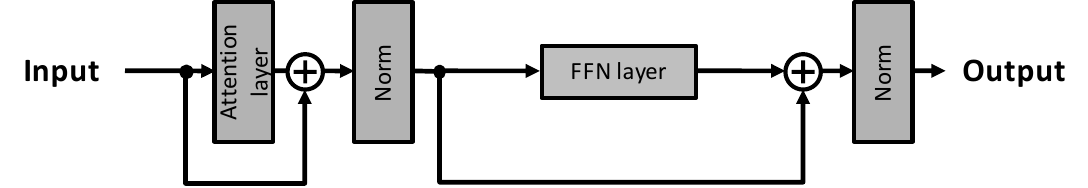}}%
\vspace{1em}
\subfloat[Sparse MoE block.]{%
\label{fig:sparse}%
\includegraphics[width=0.485\textwidth]{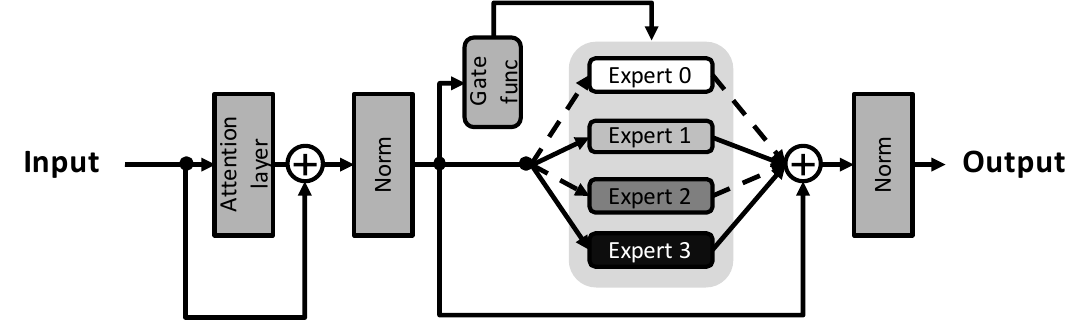}}%
\caption{
(a) A dense transformer block that consists of the self-attention layer, feed-forward networks (FFN) layer, normalizations, and residual connections. (b) An MoE block that replaces a conventional transformer block's FFN layer to induce sparsity. The example assumes the MoE block has four expert layers. Each expert has the same dimension as the FFN layer of the corresponding dense transformer block.  
}
\label{fig:dense_vs_sparse}
\end{figure}

\subsection{Sparse LLMs using Mixture-of-Experts (MoE)}
\label{sect:sparse_llm}

\textbf{MoE model architecture.}
To address the high computational requirements of dense LLMs, the Mixture-of-Experts (MoE)~\cite{moe, switchtransformer, nllbmoe, glam, gemini_1.5} model was introduced which exploits \emph{sparsity} in the model architecture to reduce LLM's high computation cost.
MoE is designed to mimic the behavior of the human brain, which consists of specialized regions that are tailored for specific tasks.
By sparsely activating only a subset of the parameters, MoE is able to scale up the model size without a corresponding increase in its computation cost (FLOPs). 

\fig{fig:sparse} illustrates the model architecture of an MoE block, which is converted from the dense transformer block in \fig{fig:dense} by replacing the original FFN layer in the transformer block with an MoE block. 
The MoE block consists of two key components: the gate function and the expert layer.
The gate function is responsible for determining the relevance of each expert for a given input token, thereby assigning probabilities to each expert based on their importance to the specific input.
The expert layer, on the other hand, is a dense FFN layer that focuses on processing distinct patterns in the input data.
During model inference, the gate function \emph{selects}  which experts should be activated for each input token based on their assigned probabilities. 
Subsequently, the activated experts process the input tokens by \emph{executing} the assigned input tokens and generate the output tokens. As such, the evaluation of an MoE block involves a two-stage process, (1) \emph{expert selection} and (2) \emph{expert execution}, an input data-dependent procedure that must be executed sequentially. 

In state-of-the-art MoE models, the number of experts that are activated is generally very small (e.g., Google's SwitchTransformer~\cite{switchtransformer} and Meta's NLLB-MoE~\cite{nllbmoe} only activates the top-1 and top-2 experts, respectively), rendering MoE's inference to exhibit high sparsity.

\begin{figure}[t!] \centering
\includegraphics[width=0.48\textwidth]{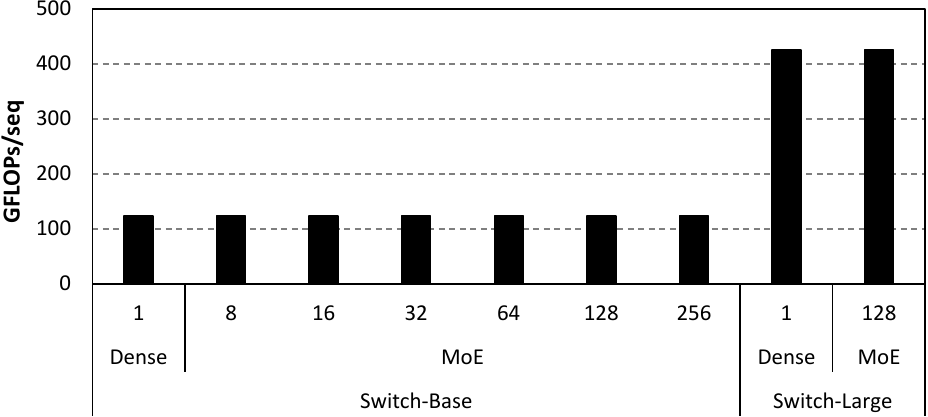}
\caption{
Required number of FLOPs per sequence in deploying
SwitchTransformer (MoE) and T5 (dense). In this figure, we show both the ``Base'' and ``Large'' model versions of SwitchTransformer and its FLOPs-equivalent T5 (\sect{sect:methodology} details the model configurations studied in this work). The numbers represent how many experts are available within the MoE block (i.e., dense T5 is equivalent to having just a \emph{single} expert). 
}
\label{fig:switch_compute}
\vspace{-0.4em}
\end{figure}

{\bf Computation cost of sparse MoE vs. dense LLMs.}
MoE's compute efficiency is achieved by selectively activating a small subset of the experts for each input token, instead of densely connecting all layers and neurons in the model.
\fig{fig:switch_compute} compares the required number of FLOPs per sequence between a representative sparse MoE model (Google's SwitchTransformer~\cite{switchtransformer}) and its dense model counterpart with an iso-FLOPs count (Google's T5~\cite{t5}).  As shown, the computation cost of MoE remains constant, regardless of the number of experts (i.e., the model size), highlighting the fact that MoE can scale up the model's capacity with minimal computation overheads.

\section{Motivation}
\label{sect:motivation}

\subsection{Key Challenges of MoE inference}
\label{sect:moe_inference}
While MoE offers advantages in scaling the LLM model size 
without significantly increasing its computation cost, it introduces several key challenges as summarized below.

\begin{enumerate}
    \item {\bf Large memory footprint.} The biggest advantage of MoE is its high compute efficiency, which comes from its ability to cost-effectively scale the model capacity by employing a large number of experts. This, however, comes at the cost of high memory consumption, leading MoE's overall memory footprint to become an order of magnitude larger than its dense counterpart, e.g., SwitchTransformer can consume as much as $75\times$ higher memory consumption than the dense T5 (\fig{fig:switch_memory}). Such large memory usage poses several obstacles for MoE, one critical challenge being its inability to fit the model within a single GPU's local memory which is only several tens of GBs in size. 

    \item {\bf Dynamic and sparse expert activation.} Because it is challenging to store the entire MoE model parameters within a single GPU, multi-GPU solutions that split the expert parameters across multiple GPU's memory can be a viable solution for high-performance MoE inference.
    Unfortunately, because MoE only partially activates the experts in a sparse manner, the number of experts actually executed by each GPU for inference becomes very low. In effect, multi-GPU solutions for deploying MoE suffer from low GPU compute utilization and deteriorate the TCO.
    Furthermore, because MoE experts are sparsely activated, a significant fraction of expert parameters allocated inside the expensive GPU memory is most likely not going to be utilized when servicing any given inference request (e.g., SwitchTransformer activating top-1 among the 128 experts executes only $0.8$\% of its experts per inference).
    Finally, because the sparsely activated experts are determined dynamically in an input data dependent manner, it becomes challenging to predict which experts will be activated at runtime, preventing any load-balancing solutions to better distribute the number of activated experts across the GPUs in an even manner.
    Since GPU's limited memory capacity was the very reason why a multi-GPU system was necessary for deploying MoE, such sub-optimal utilization of GPU memory is a significant waste.

\end{enumerate}

\begin{figure}[t!] \centering
\includegraphics[width=0.48\textwidth]{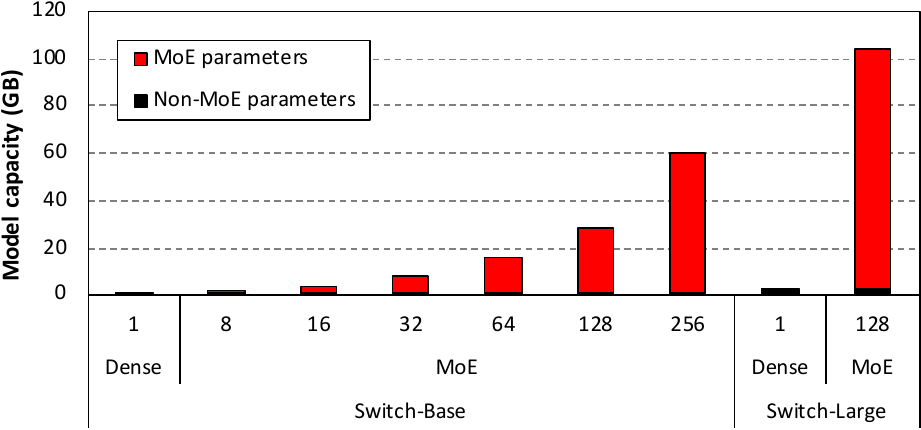}
\caption{
Memory capacity requirement of deploying
SwitchTransformer (MoE) and T5 (dense). 
 MoE parameters include both the expert layer and the gate function, while the rest of the layers are marked as Non-MoE parameters. As depicted, MoE expert parameters account for the majority of the model's memory consumption.
}
\label{fig:switch_memory}
\vspace{-0.4em}\end{figure}

\subsection{Prior Solution: CPU Offloading of Expert Parameters}
\label{sect:offloading}

The challenge of efficiently managing LLM's large model size within the constraints of limited GPU memory has led to several prior work advocating to offload the memory hungry LLM parameters to CPU DRAM or even SSD~\cite{zero_offload, zero_infinity}. These \emph{CPU offload} based approaches have also been explored under the context of MoE in order to address its aforementioned challenges, i.e., its large memory footprint and high deployment cost~\cite{huggingface_accelerate, dsinf, semoe, meta_moe}. Although  CPU offload based solutions can help reduce the number of GPUs required for servicing MoE, the latency to transfer the CPU offloaded model parameters to the GPU memory can deteriorate end-to-end performance. This is because none of the prior work fundamentally addresses the dynamic and sparse expert activation challenge and its sequential dependency issue. Below we classify prior CPU offload based solutions into two categories, (1) \emph{fetch-on-demand} and (2) \emph{prefetch-all}, discussing its benefits as well as its limitations. 

\textbf{Fetch-on-demand.} This design point~\cite{huggingface_accelerate} employs the fetch-on-demand based CPU offloading for MoE serving. Under this system design, \emph{all} the expert parameters are offloaded to the capacity-optimized CPU memory.
At runtime, once the expert selection stage identifies which experts are activated, those activated experts are migrated to the GPU memory \emph{on-demand}.
Because GPU memory is only used to store the activated experts (and not the entire experts as done in a baseline multi-GPU system), it helps improve GPU memory utilization significantly.  
However, the process of migrating activated experts on-demand serializes the expert selection stage with the expert execution stage, causing noticeable performance overhead. In the rest of this paper, we refer to this design point as \emph{\ondemand}.

\textbf{Prefetch-all.} To better hide the CPU$\rightarrow$GPU expert transfer latency, prior work on SE-MoE~\cite{semoe} proposes a \emph{prefetching} based CPU offloading for MoE, where the expert parameters are proactively migrated to the GPU memory before its actual usage (henceforth referred to as \emph{\prefetch}). Similar to \ondemand, \prefetch offloads all expert parameters in CPU memory. 
\prefetch then migrates the \emph{entire} expert parameters to be used by the \emph{next} MoE block while the \emph{current} MoE block's expert execution is taking place. While \prefetch can help overlap compute (current MoE block's expert execution) with communication (transferring all experts required for the next MoE block), it suffers from several limitations. First, \prefetch is not scalable as CPU$\rightarrow$GPU expert transfer time can be prohibitive when there exists a large number of experts 
(e.g., Google’s SwitchTransformer~\cite{switchtransformer} contains up to 256 experts, while Meta’s NLLB-MoE~\cite{nllbmoe} employs 128 experts).
Second, at any given MoE block's execution, the GPU memory must be large enough to store both the current as well as the next MoE block's entire expert layer parameters, potentially overwhelming the scarce GPU memory.
To mitigate the significant GPU memory demands required for transferring entire expert parameters, one could consider prefetching only a subset of experts predicted to be active in the next block. Nonetheless, predicting active experts does not always ensure accuracy, and mispredictions can lead to penalties such as waste of GPU memory and the unavoidable serialization of expert transfer latency, consequently increasing the end-to-end inference latency.

Aside from these prior works focusing on CPU offloading decisions for MoE inference, 
\cite{meta_moe} characterizes MoE deployment's inference latency and its memory usage across different components of the MoE model architecture, suggesting several optimization strategies like dynamic gating and expert buffering. Dynamic gating helps reduce wasted memory allocations for multi-GPU MoE inference and expert buffering is a technique that can help reduce inference latency by caching hot, active experts in GPU memory. In \sect{sect:eval:discussion}, we further discuss the applicability of expert caching on top of MoE-offload design points.

Overall, we conclude that the inherent sequential dependency between the expert selection and expert execution stage poses several challenges in designing a performance-efficient CPU offloading based MoE system. The key objective of this paper is to exploit the dynamic and sparse nature of MoE models to design a holistic system solution that effectively balances memory efficiency and high performance. 

\section{\proposed: Co-Designing Algorithm and System for Fast $\&$ Scalable MoE Inference} 
\label{sect:proposed}

\subsection{High-level Overview}
\label{sect:high_level_overview}

\begin{figure}[t!] \centering
\includegraphics[width=0.4\textwidth]{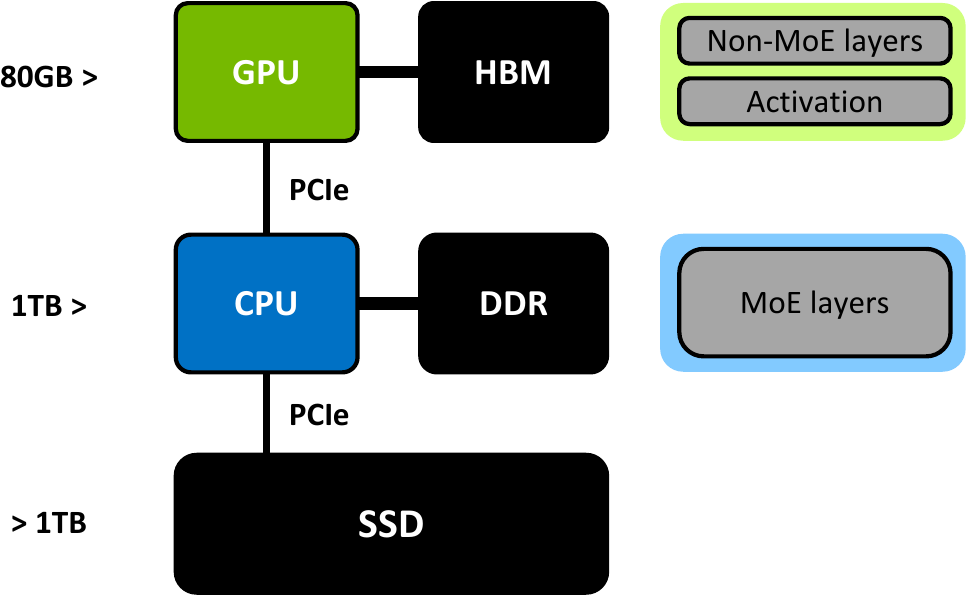}
\vspace{0.7em}
\caption{
\proposed system for deploying MoE-based LLMs. The memory hungry, sparse MoE parameters are all offloaded to the capacity-optimized CPU memory and are only transferred to the GPU memory when necessary for inference. The rest of the dense, non-MoE parameters are stored locally within the GPU memory.
}
\label{fig:deployment}
\vspace{-0.4em}\end{figure}

We propose \proposed, an algorithm-system co-design for scalable and high-performance MoE inference. \proposed is designed to address the large memory footprint challenge of MoE while also mitigating the dynamic nature of sparse expert activation for performance improvement. These features enable our \proposed to deploy large-scale LLM using just a single GPU.
Because activated experts are determined dynamically in an input dependent manner, all expert parameters must be preserved at all times, regardless of its actual utilization.
To efficiently manage the storage of MoE's substantial model capacity, \proposed carefully considers the model parameter's actual utility to decide their storage locations. As shown in \fig{fig:deployment}, the dense non-MoE parameters are stored locally within the GPU memory as they are always utilized, regardless of the input values.
Meanwhile, the sparse MoE parameters are completely offloaded to the CPU's DRAM because (1) they account for the majority of LLM's model capacity so CPU offloading can help significantly save GPU memory, and (2) only a small fraction of the MoE experts that are activated are actually utilized for inference.
As we detail in this section, such hierarchical storage of MoE parameters and its deployment proves effective in minimizing the usage of GPU memory while still providing high performance.

As discussed in \sect{sect:offloading}, prior work has followed two main approaches, \ondemand and \prefetch.
Because traditional MoE blocks must sequentially execute expert selection followed by expert execution in an input data dependent manner, both \ondemand and \prefetch suffer from sub-optimal performance. In particular, \ondemand directly exposes the CPU$\rightarrow$GPU communication latency to migrate activated experts as part of end-to-end inference time. This is because expert execution must always be preceded with the expert selection stage.
\prefetch can hide the communication latency to transfer expert parameters to some extent, but it still suffers from performance loss because \emph{all} expert parameters must be transferred to the GPU, even though only a small fraction of them will actually be utilized for inference.

The key objective of  \proposed is to mitigate the impact of the MoE block's dynamically determined sparse expert activation and utilize that property for performance improvement. Specifically, \proposed introduces a new gate function that
\emph{decouples} the expert selection stage from the expert execution stage. The benefit of decoupling expert selection with expert execution is twofold. First, it enables our system to significantly reduce the latency to migrate experts from CPU to GPU as only the activated experts will be migrated under our proposed design. 
Second, the performance overhead of migrating the activated experts can be effectively hidden by overlapping it with MoE block’s computation.
In the remainder of this section, we detail the two key facets of our algorithm-system co-design.

\begin{figure}[t!] \centering
\includegraphics[width=0.485\textwidth]{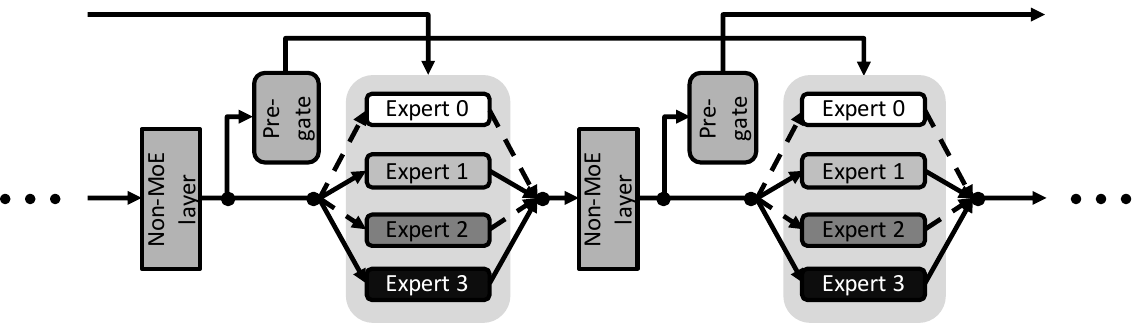}
\caption{
Two consecutive MoE blocks employing our proposed pre-gate function. For brevity, we only provide a detailed illustration on the MoE blocks (the rest of the non-MoE layers are consolidated into a single block in this figure). The residual paths within a transformer block are not shown.
As depicted, a pre-gate function is trained to select which experts to activate for the next MoE block.
}
\label{fig:pregate}
\vspace{-0.4em}
\end{figure}

\subsection{(Algorithm) Pre-gated MoE Architecture}
\label{subsect:model_arc}

{\bf Pre-gate function.} In traditional MoE model architectures, each MoE block contains a gate function which selects the experts to activate within the \emph{same} MoE block. Because only those experts that are activated are subject to the subsequent expert execution stage, it is impossible to overlap the expert selection stage with the expert execution stage. In our proposed MoE model architecture, we introduce the \emph{pre-gate function} which is trained to preemptively select the experts to activate for the \emph{next} MoE block rather than the current MoE block. More concretely, a pre-gate function for the $N$-th MoE block is trained to generate the activation masks to utilize in the ($N$+1)-th MoE block to
select which experts to activate (\fig{fig:pregate}).
Prior work has explored alternative ways to train gate functions, which are fine-tuned for specific objective functions such as enhancing the model accuracy and alleviating the input token's load-imbalance problem when distributed across multiple GPUs~\cite{choicerouting, baselayer, switchtransformer, hashlayer, thor}. 
\old{The approach taken with our Pre-gated MoE is aligned with these prior art but with one important distinction -- our pre-gate function is designed to gain foreknowledge of which experts to select for the subsequent MoE block.}
The approach taken with our Pre-gated MoE is aligned with these prior art but with one important distinction -- our pre-gate function is designed to deterministically select and pre-compute which experts to activate for the subsequent MoE block.
As we demonstrate in \mbox{\sect{sect:eval:accuracy}}, our pre-gate function has a minimal impact on LLM's model accuracy and is shown to be highly robust.

\begin{figure}[t!] \centering
\includegraphics[width=0.485\textwidth]{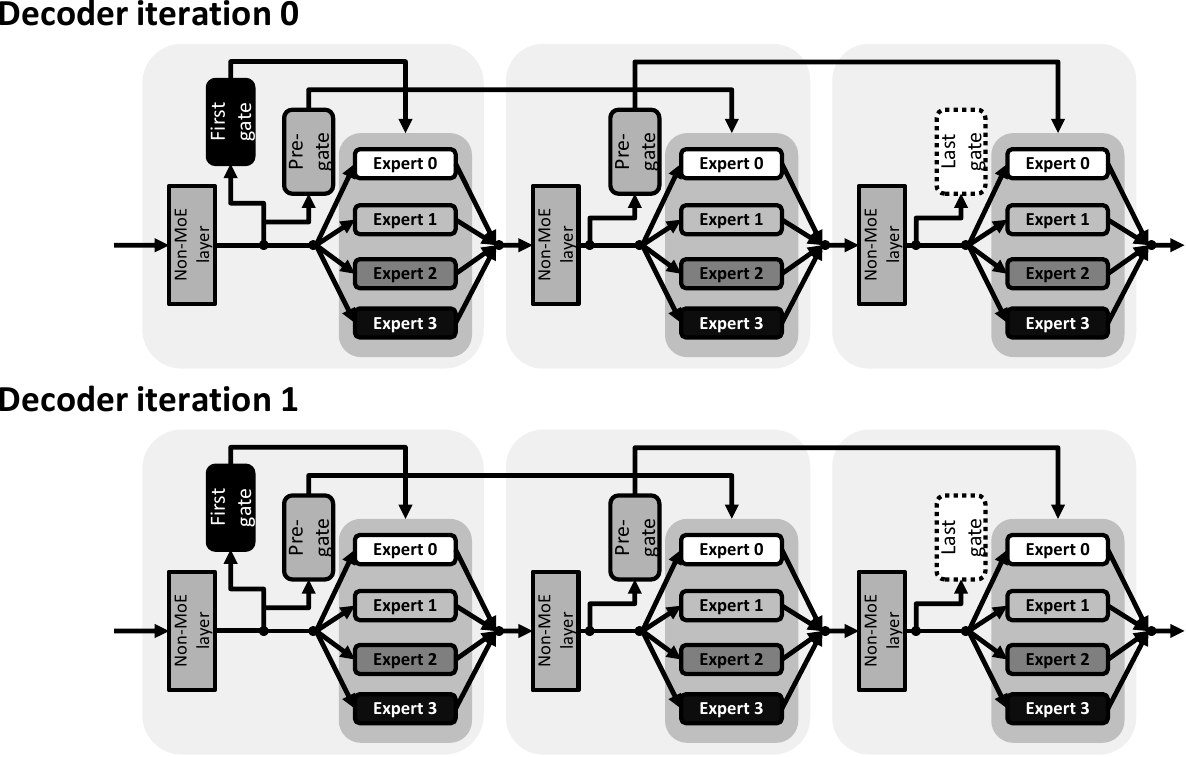}
\caption{
The sequence of pre-gated MoE block's execution during the course of two consecutive decoder iterations. We assume the LLM consists of three pre-gated MoE blocks, requiring three MoE block executions for a single iteration of decoding. The first MoE block employs two gate functions (one for the current MoE block and another for the next MoE block) whereas the last MoE block does not utilize any gate function.
}
\vspace{-0.4em}
\label{fig9_pregate_moe_decoder}
\end{figure}

Since our pre-gate function is trained to select the active mask for the next MoE block, two important questions remain: (1) How does our Pre-gated MoE architecture select the experts to activate for the \emph{first} MoE block (i.e., the first MoE block does not have a previous MoE block that will select the experts to activate on behalf of the first block)? (2) What is the role of the pre-gate function for the \emph{last} MoE block (i.e., the last MoE block does not have a subsequent MoE
block)? We answer these questions using \fig{fig9_pregate_moe_decoder}. In conventional MoE-based LLMs, a single decoder iteration generates a single output token (word) and multiple iterations of decoding are conducted during a single inference run to generate the final output result (which is a series of tokens). As shown in \fig{fig9_pregate_moe_decoder}, a single decoder iteration involves the execution of several stacks of MoE blocks. In our proposed MoE design, the first MoE block employs \emph{two} gate functions, the first gate selecting the activated experts for the first MoE block (identical to conventional MoE architectures) and the second gate (our \emph{pre}-gate function) selecting the experts to activate for the second MoE block. Conversely, because the last MoE block does not have a subsequent MoE block to execute within the \emph{same} decoder iteration, we do not employ a pre-gate function for the last MoE block. In effect, the pre-gate function does not select activated experts \emph{across} different decoder iterations.

{\bf Training the pre-gate function.}
Today's LLMs are first \emph{pretrained} on vast amounts of textual data which spans a wide variety of languages and application domains. The pretraining stage
requires a massive amount of computation power (several tens of thousands of GPUs) and typically takes several months to complete (e.g., GPT-3 is pretrained over hundreds of billions of tokens for more than one month using thousands of GPUs~\cite{gpt3}).
Once the LLM is pre-trained, it goes through the \emph{fine-tuning} stage with task-specific datasets for specific use cases (e.g., summarization, question answering).
Training our Pre-gated MoE does not change how the resource-intensive pretraining stage is conducted, as our pre-gate functions are incrementally trained during the fine-tuning stage.
Specifically, we utilize existing pretrained MoE model parameters as-is but change the MoE model architecture to properly accommodate the functionalities of our pre-gate function as well as the augmentations required in the first/last MoE block (see \fig{fig9_pregate_moe_decoder}). We then go through the fine-tuning stage as required by the downstream task, identical to how conventional MoE models will be fine-tuned in a task specific manner. When our Pre-gated MoE is fine-tuned over the same number of fine-tuning training iterations vs. conventional MoE models, we observe no noticeable degradation in LLM model accuracy, one which we further elaborate in \sect{sect:eval:accuracy}.

\begin{figure}[t!] \centering
\includegraphics[width=0.485\textwidth]{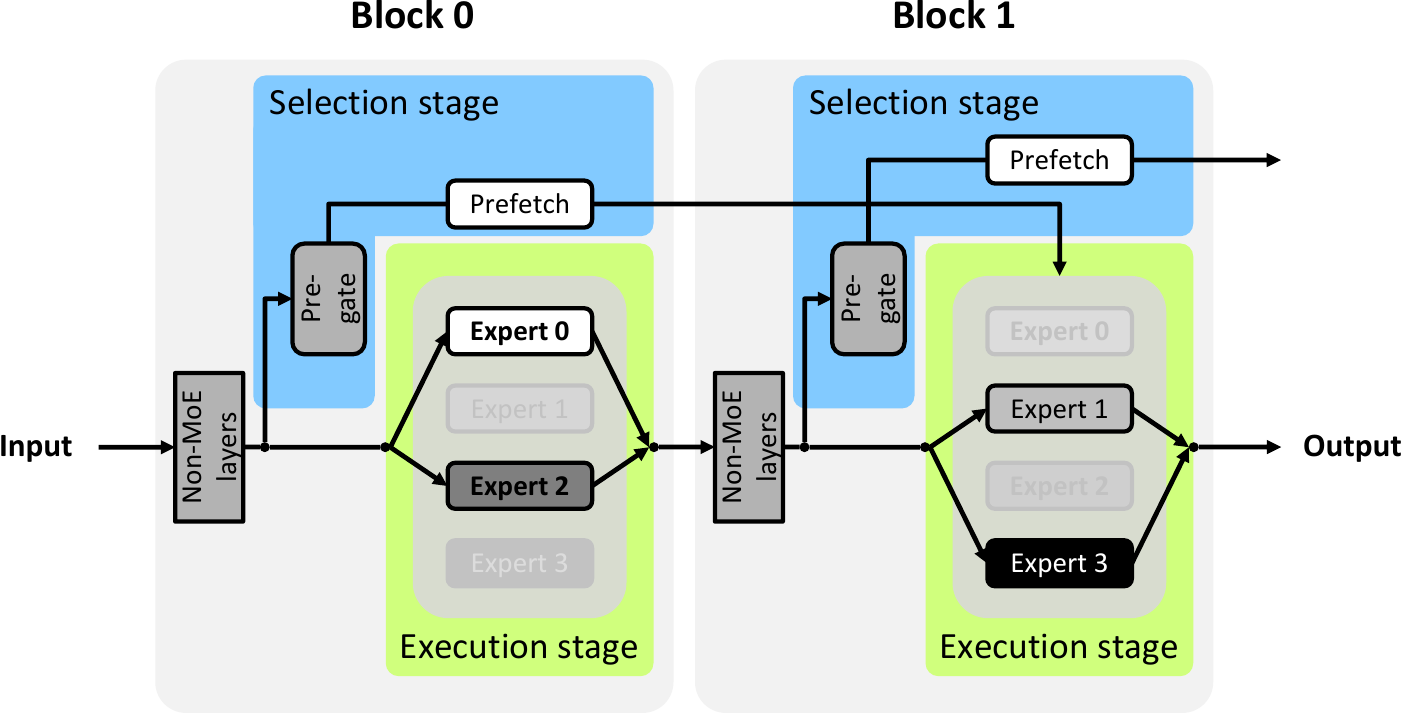}
\caption{
Our pre-gate function helps eliminate the sequential dependency between an MoE block's expert selection and expert execution stage. The gate function is implemented as a compact MLP layer having low computation requirement, so the preemptive migration of activated experts right after the gate function (blue) exhibits a PCIe communication-bound behavior. Overall, our Pre-gated MoE enables the compute-bound expert execution stage (green) to concurrently execute with the communication-bound expert selection stage (blue) for all MoE blocks (with the exception of the first MoE block). Example assumes that expert $0$ and $2$ are activated for the first MoE block while expert $1$ and $3$ are activated for the second MoE block.
}
\label{fig:stage}
\vspace{-0.4em}
\end{figure}

\subsection{(System) Preemptive Expert Migration}
\label{subsect:system_arc}

Our pre-gate function provides MoE models with the ability to determine what experts will be activated in the next MoE block while the current MoE block is being executed, presenting new opportunities for system-level performance optimizations. In particular, with the exception of the first MoE block, all MoE block's expert execution stage is now completely decoupled from the expert selection stage without any data dependencies, allowing both stages to be concurrently executed (\fig{fig:stage})\footnote{The first MoE block is the only exception to this property under our Pre-gated MoE design -- due to the lack of a pre-gate function in the first MoE block, we must sequentially execute its expert selection and expert execution stages, identical to conventional MoE models. Because state-of-the-art LLMs typically contain tens of MoE blocks, most of the MoE blocks are able to overlap expert selection with expert execution.}. Such feature opens up several opportunities as detailed below.

\begin{figure}[t!] \centering
\includegraphics[width=0.485\textwidth]{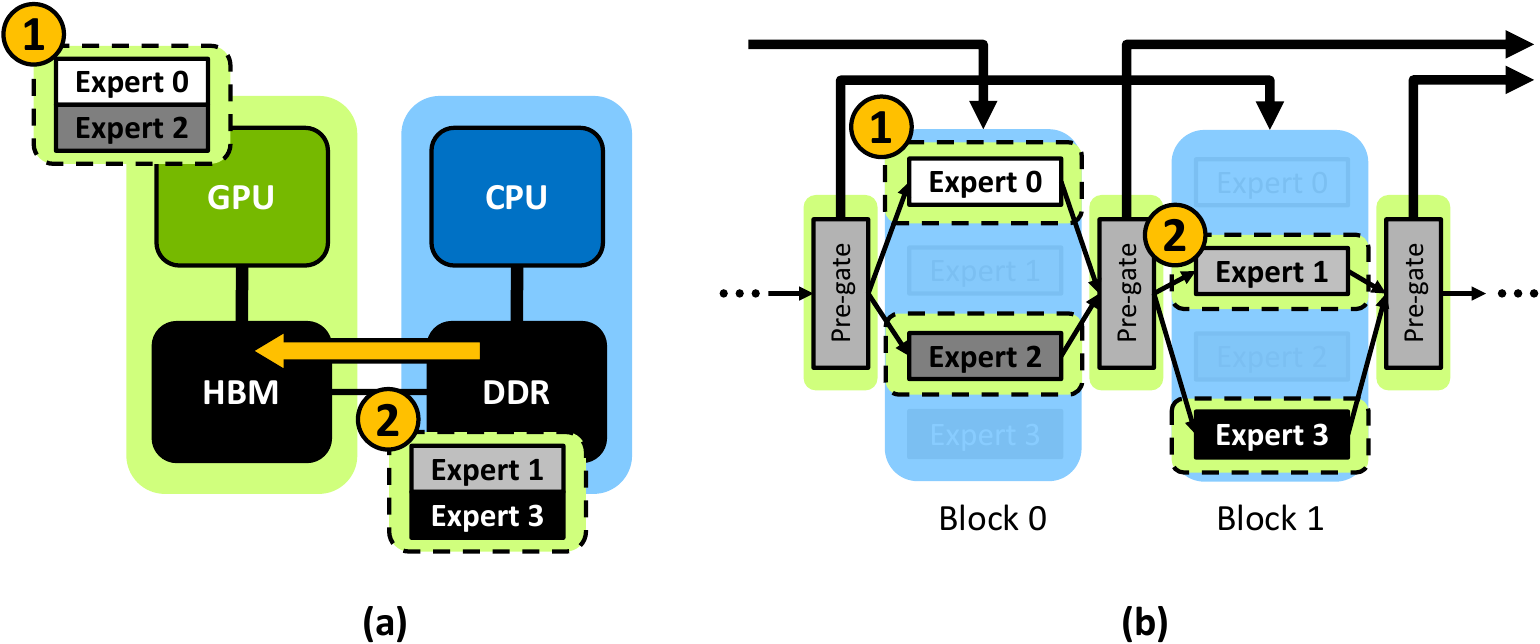}
\caption{(a) Illustration of how the example in \fig{fig:stage} gets handled over the Pre-gated MoE system where the expert execution in MoE block $0$ (using the activated experts $0$ and $2$) concurrently takes places while the next MoE block $1$'s activated experts $1$ and $3$ are migrated over to the GPU memory. (b) The on-demand migration of just the activated experts (green) allows the entire experts to be stored in CPU memory (blue), significantly saving GPU memory capacity.
} 
\label{fig:system}
\vspace{-0.4em}
\end{figure}

 {\bf CPU offloading with minimal expert migration overhead.} A key limitation of previous CPU offloading solutions is that the latency to migrate the CPU-offloaded expert parameters is either directly exposed as part of the end-to-end inference time (\ondemand) or the size of the migrated experts are simply too large that, despite its opportunity to overlap expert migration with the expert execution, copying the experts overwhelms the end-to-end performance (\prefetch). Our Pre-gated MoE, on the other hand, can evaluate which experts will be activated in advance, only migrating the activated experts for the next MoE block while the current MoE block's experts are being executed. 
This effectively addresses the dual challenges of prior CPU offloading techniques, namely (a) \ondemand's serialization of expert selection and expert execution (resolved by Pre-gated MoE's concurrent expert migration and expert execution) and (b) \prefetch's large expert migration latency (tackled by Pre-gated MoE's ability to only migrate activated experts).
State-of-the-art MoE models employ a large number of experts within an MoE block while only activating a very small subset of them (e.g., Google's SwitchTransformer~\cite{switchtransformer} contains up to $256$ experts but only activates the top-1 expert, while Meta's NLLB-MoE~\cite{nllbmoe} employs $128$ experts and activates top-2 experts). As depicted in \fig{fig:system}, we can clearly see the benefit of how our algorithm-system codesign can effectively address the limitations of existing CPU-offloading solutions, maximizing the opportunity to overlap expert migration latency with expert execution time while also ensuring that the CPU$\rightarrow$GPU data transfer size is minimized.
\fig{fig:timeline} points out the limitations of \ondemand and \prefetch and how our Pre-gated MoE successfully addresses its shortcomings, potentially reaching the performance of an ideal, \emph{GPU-only} design point when the latency to migrate activated experts can be completely hidden inside the MoE expert execution stage.

\begin{equation}
\begin{split}\label{eqn:peakgpumem}
\forall\ N, \quad 0 &\leq N < Number\ of\ MoE\ blocks\\
Peak\_GPU\_mem &= \max\left(Non\_MoE_{M} + \sum_{L=N}^{N+1} Act\_Exp_{L}\right) 
\vspace{0.8em}
\end{split}
\end{equation}

{\bf Low GPU memory utilization for large LLM deployment.}
The majority of MoE-based LLM's model capacity are concentrated around MoE parameters. Since our Pre-gated MoE system offloads the entire MoE parameters to CPU memory and only migrates activated experts over to the GPU, we are able to significantly reduce the \emph{peak} usage of GPU memory. 
In our \proposed system, peak GPU memory usage is primarily dominated by the memory capacity required to store (1) all the non-MoE parameters (which are statically stored in GPU memory) and (2) the active experts for both the current and the subsequent MoE block (dynamically determined at runtime and copied over to the GPU memory). \eqn{eqn:peakgpumem} summarizes the peak GPU memory usage to store MoE-based LLM's model parameters under \proposed.

\begin{figure}[t!] \centering
\includegraphics[width=0.485\textwidth]{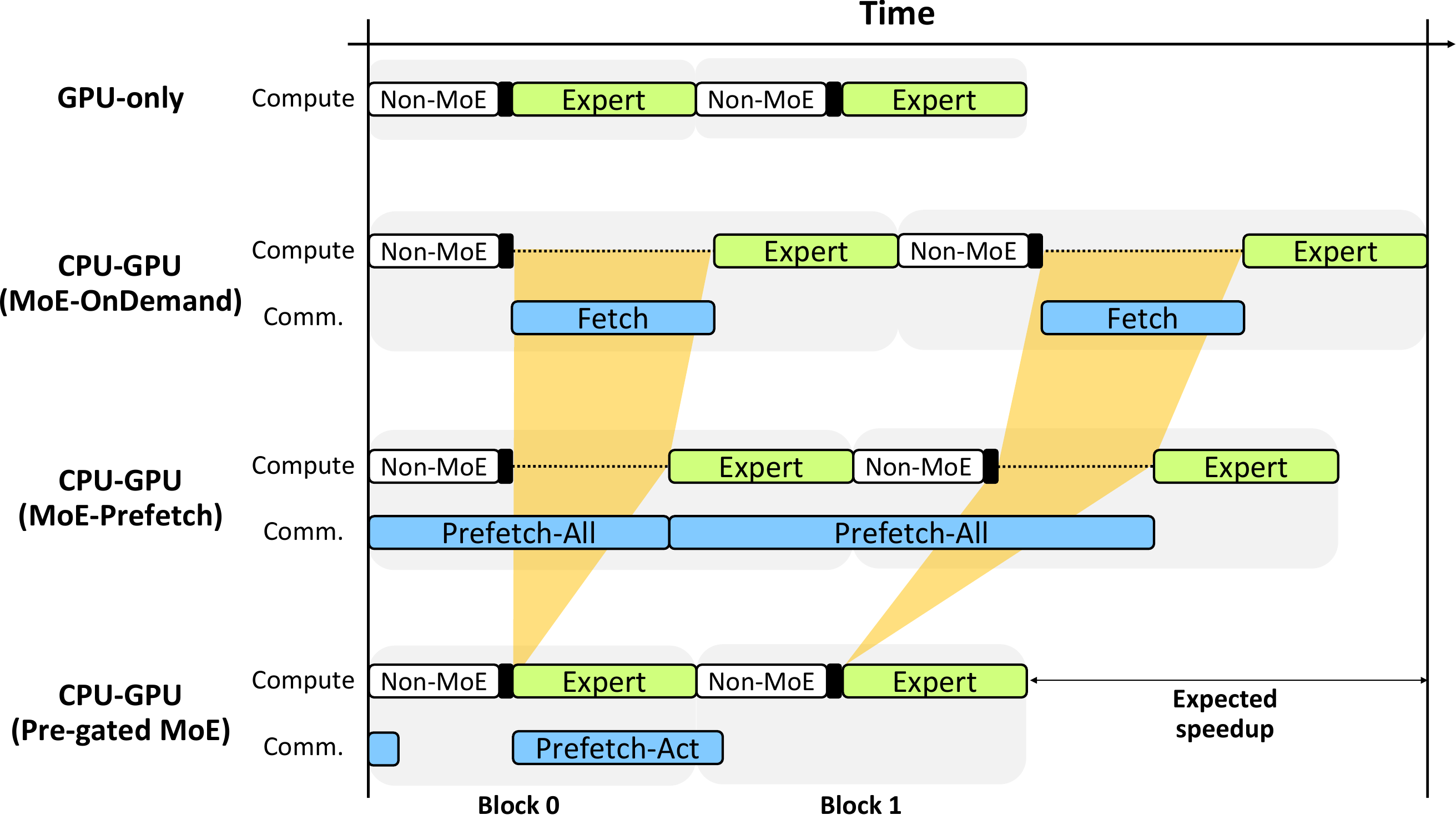}
\caption{
Execution timeline between our \proposed system and three baseline designs (GPU-only, \ondemand, and \prefetch). The black bar represents the latency to execute the gate functions. GPU-only is an ideal, oracular design point that has infinite GPU memory capacity, allowing the entire model parameters to be stored within GPU memory (i.e., there is no communication latency to migrate experts from CPU to GPU).
In our Pre-gated MoE, the latency to migrate experts can be hidden by both the expert and non-MoE layer's (e.g., self-attention layer) execution time.
} 
\label{fig:timeline}
\vspace{-0.4em}
\end{figure}

In this equation, \(Non\_MoE_{M}\) represents the total size of the non-MoE parameters while $\sum_{L=N}^{N+1} Act\_Exp_{L}$ represents the aggregate size of the active expert parameters over two consecutive (the $N$-th and ($N$+1)-th) MoE blocks. Since expert parameters account for the majority of MoE-based LLM's model size (see \fig{fig:switch_memory}) and only a small fraction of experts are activated during inference, the peak GPU memory usage in \eqn{eqn:peakgpumem} becomes much lower than \gpuonly and can also reach the memory consumption level of the memory-optimal \ondemand design. A key advantage of reducing peak GPU memory usage is that it facilitates the deployment of considerably larger LLMs on systems with limited GPU memory resources (e.g., desktop and edge devices).
In \sect{sect:eval:scalability}, we demonstrate Pre-gated MoE's scalability and applicability for deploying large-scale LLMs.

\section{Methodology}
\label{sect:methodology}

{\bf System configuration.}
We conducted our evaluation using two system design points, \gpuonly and \gpucpu, which utilize an AMD EPYC 7V12 64-Core CPU with  1.8TB DDR4 memory and a single NVIDIA GPU A100 with 80GB of HBM. The CPU and GPU communicate over a PCIe (gen4) channel with $32$ GB/sec of data transfer bandwidth.

The oracular \gpuonly design assumes the entire model parameters are stored in GPU memory, so the all computations for inference are conducted on the GPU. 
Note that multi-GPU solutions leveraging expert parallelism can experience performance loss due to inter-GPU communications and load imbalance issues. For a conservative evaluation, we experiment with our \gpuonly system under a single GPU system that can achieve the highest performance.
As such, \gpuonly represents a performance-optimal, upper-bound MoE inference system that we compare our \proposed against.

The \gpucpu design, on the other hand, utilizes both GPU and CPU memory for storing the model parameters where only the (dense) non-MoE parameters are persistently stored within the GPU memory while the (sparse) MoE parameters are completely offloaded to CPU memory (\fig{fig:deployment}). 
Our \proposed system as well as the two baseline CPU offloading MoE systems (\ondemand and \prefetch) employ such \gpucpu system configuration.

\begin{table}[t]
    \centering
    \hfill
    \caption{Model configuration of Google's SwitchTransformer.}
\scriptsize
    \begin{tabular}{|c||c|c|c|c|}
    \hline
    \textbf{Model} & \textbf{Experts} & \textbf{Layers} & \textbf{Parameters (B)} & \textbf{Capacity (GB)} \\
    \hline
    \multirow{3}{*}{Switch-Base}
            & 8 & 12 & 0.7 & 2.8 \\
            \cline{2-5}
            & 64 & 12 & 3.8 & 15.2 \\
            \cline{2-5}
            & 128 & 12 & 7.5 & 30.0 \\
    \hline  
    \multirow{1}{*}{Switch-Large}
            & 128 & 24 & 26.4 & 105.6 \\
    \hline
    \end{tabular}

    \label{tab:model_config}
\end{table}

{\bf Model and dataset.}
We use Google's SwitchTransformer~\cite{switchtransformer}
as the baseline MoE for our evaluations, a state-of-the-art large-scale MoE model.
The open-sourced pretrained weights available at HuggingFace~\cite{huggingface} were utilized to fine-tune both \proposed as well as the baseline MoE model for downstream tasks (\tab{tab:model_config}).
As for the training data, we study three datasets covering two distinct downstream tasks: one from the summarization task (Xsum~\cite{xsum}) and two from the closed-book question answering task (CB Web QA~\cite{cbwebqa}, SQuAD~\cite{squad}).
The evaluation metrics included Rouge-1 and Rouge-2 scores~\cite{rouge} for summarization, and ExactMatch and F1 scores for question answering.

\textbf{Model training (fine-tuning).}
We applied the exact same fine-tuning configurations across all model architectures including \proposed and conventional MoE. As discussed in \sect{subsect:model_arc}, the fine-tuning stage utilizes the pre-trained weights from the conventional MoE model. We utilize a mini-batch containing 256 sequences, each with a length of 256 tokens, to fine-tune the model for 2,048 steps (i.e., $2^{27}$ tokens in aggregate). A constant learning rate of 0.0001 is employed.

{\bf Software implementation.}
All of our GPU-only and CPU-GPU systems are implemented using NVIDIA's FasterTransformer~\cite{fastertransformer}, a state-of-the-art high-performance CUDA library widely employed in production inference servers in the industry.
Because end-to-end inference performance is less sensitive to what the downstream task the MoE model is trained for, we report performance numbers using the MoE model fine-tuned for the closed-book question answering tasks with the SQuAD dataset. When reporting model accuracy, we use the two downstream tasks as discussed above.

\section{Evaluation} 
\label{sect:eval}

In this section, we first demonstrate \proposed's effectiveness in improving performance (\sect{sect:eval:perf}) and discuss its scalability to large-scale MoE models (\sect{sect:eval:scalability}). We then quantitatively evaluate \proposed's impact on model accuracy (\sect{sect:eval:accuracy}) and finally present sensitivity studies as a discussion point (\sect{sect:eval:discussion}), demonstrating the robustness of \proposed.

\begin{figure}[t!] 
\includegraphics[width=0.49\textwidth]{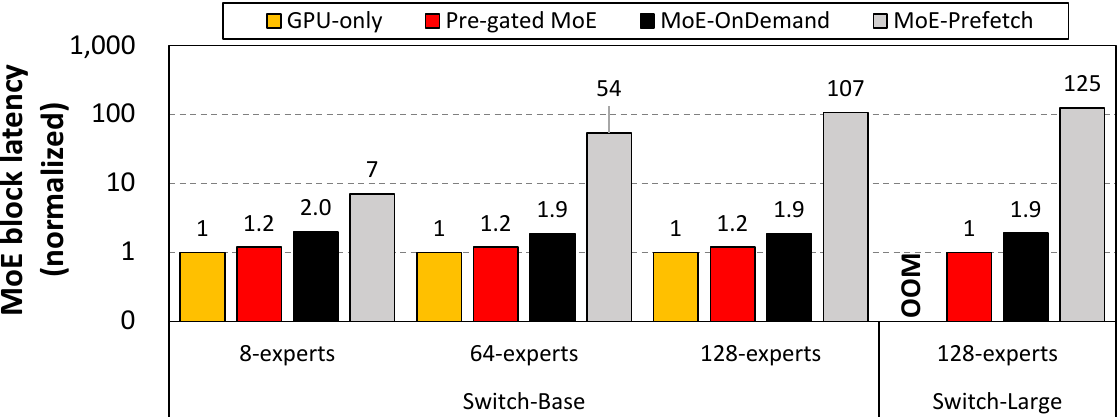}
\caption{Average latency incurred in executing a single MoE block (normalized to \gpuonly). Since \gpuonly experiences an out-of-memory (OOM) error in Switch-Large, we normalized the latency of \ondemand and \prefetch to \proposed.
Note that the y-axis in this chart is plotted in log-scale.
} 
\vspace{-0.4em}
\label{fig:moe_block_latency_model}
\end{figure}

\subsection{Performance}
\label{sect:eval:perf}

In this section, we primarily focus on single batch inference scenarios because real-world production ML serving systems are optimized for a batch size of $1$~\cite{brainwave_isca, mlperf_inf, optq}. As discussed in \sect{subsect:model_arc},   the end-to-end performance of CPU offloading solutions are primarily determined by how well the CPU$\rightarrow$GPU communication time (to migrate experts) is hidden inside the MoE block's execution time. Furthermore, the end-to-end MoE inference time is mostly dominated by a series of (identically sized) MoE block's execution. 
As such, we first focus on comparing a single MoE block's execution time between \proposed vs. baseline systems. 
We then discuss the improvements in end-to-end inference throughput, measured as the number of tokens processed per second.

{\bf MoE block latency.} 
\fig{fig:moe_block_latency_model} summarizes the average latency in executing a single MoE block. 
Across all configurations, \proposed significantly reduces latency by an average $1.7$\x (max $1.9\times$) and $42$\x (max $125\times$) vs. \ondemand and \prefetch, respectively.
\proposed also exhibits comparable latency to the performance-optimal \gpuonly, incurring only $19\%$ latency overhead across all Switch-Base model configurations.
Because \prefetch must migrate all experts, it suffers from the highest latency where the larger number of experts directly translates into higher performance overheads. \ondemand does better than \prefetch, thanks to its ability to only migrate activated experts. However, \ondemand still suffers from longer latency than \proposed due to the serialization of expert selection and expert execution stages.

It is worth pointing out that the performance-optimal \gpuonly is unable to run the largest MoE model, i.e., Switch-Large with 128 experts (105.6 GB), due to the limitations in GPU memory capacity, resulting in an out-of-memory (OOM) error.
\proposed still shows the shortest latency among the three \gpucpu based designs with Switch-Large, achieving $1.9$\x and $125$\x latency reduction than \ondemand and \prefetch, respectively.

{\bf End-to-end inference throughput.}
\fig{fig:throughput_all} shows the end-to-end inference throughput across all  model configurations.
\proposed achieves an average 111 tokens/sec throughput over all Switch-Base model configurations, an average $1.5$$\times$ (max $1.6\times$) and $27$$\times$ (max $55\times$) improvement over \ondemand and \prefetch, respectively. Furthermore, \proposed is able to achieve $81\%$ of the throughput of oracular \gpuonly solution, demonstrating its superior cost-effectiveness. As for the Switch-Large model with 128 experts, \proposed achieves 42 tokens/sec of throughput which is $1.6$\x and $52$\x higher than \ondemand and \prefetch, respectively.

\begin{figure}[t!] \centering
\includegraphics[width=0.49\textwidth]{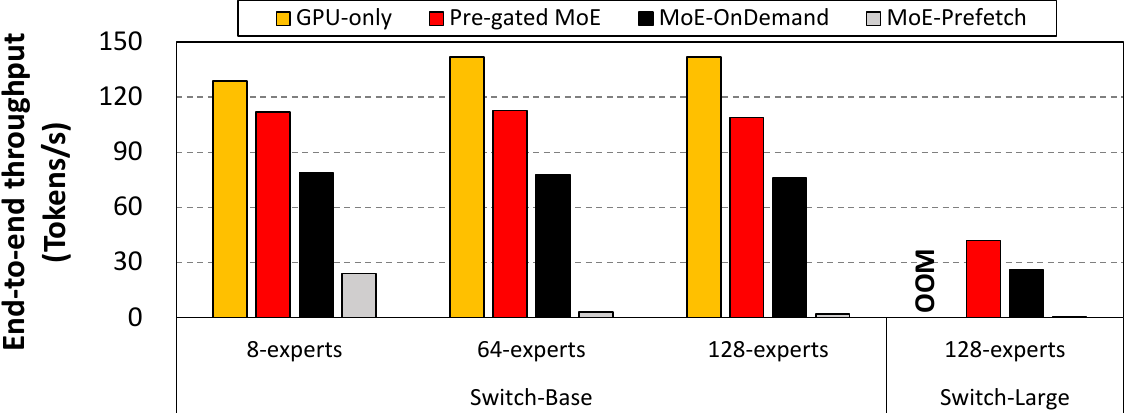}
\caption{
End-to-end inference throughput.
GPU-only experiences an out-of-memory (OOM) error in Switch-Large.
} 
\vspace{-0.5em}
\label{fig:throughput_all}
\end{figure}

\subsection{Scalability}
\label{sect:eval:scalability}

As discussed in \sect{sect:sparse_llm}, the majority of MoE's model size is dominated by expert parameters yet only a small fraction of the experts are actually activated for execution. Consequently, judiciously allocating GPU memory for efficient usage  
becomes vital in minimizing GPU's peak memory usage which helps deploy large-scale LLMs. 
\fig{fig:peak_gpu} compares the peak GPU memory usage of \proposed against baseline systems to demonstrate \proposed's scalability.

Among the four designs, \gpuonly shows the highest peak memory usage because it solely relies on GPU memory to allocate all of its model parameters and input/output activations. 
All three \gpucpu systems are able to significantly reduce peak GPU memory usage, as the memory hungry expert parameters are offloaded to the CPU memory. Also, notice how the GPU memory usage gap between \gpuonly and the three CPU offloading based \gpucpu designs gradually increases as the number of experts are increased. This is because the larger the number of experts are available within an MoE block, the more GPU memory savings the CPU offloading will provide. 
\prefetch, however, still consumes an average $51$\% of \gpuonly's peak GPU memory usage because it always migrates the entire expert parameters to GPU memory. The memory-optimal \ondemand does much better than \prefetch as it only migrates activated experts on-demand, showing the lowest peak GPU memory utilization. Our proposed \proposed system is able to consume only $23\%$ of \gpuonly's peak GPU memory usage while only incurring $0.2\%$ more GPU memory consumption vs. the memory-optimal \ondemand.

Overall, these results demonstrate that \proposed is capable of reaching the performance provided with the performance-optimal \gpuonly (\fig{fig:throughput_all}) while also achieving the resource-efficiency of the memory-optimal \ondemand, achieving high scalability to deploy large LLMs.

\begin{figure}[t!] \centering
\includegraphics[width=0.49\textwidth]{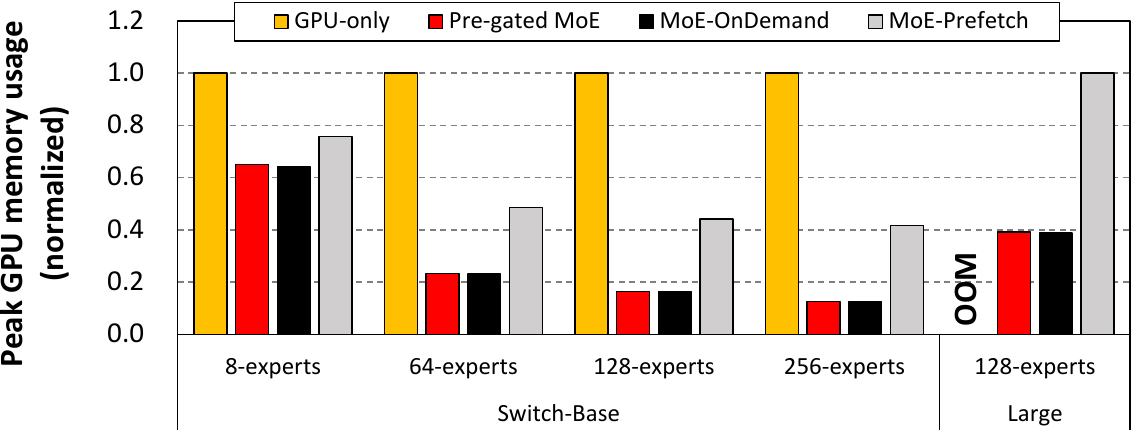}
\caption{Peak GPU memory consumption (normalized to \gpuonly). We additionally evaluate Switch-Base with $256$ experts to further demonstrate \proposed's scalability in deploying larger MoE-based LLMs. 
For the Switch-Large with 128 experts, \gpuonly suffers from an OOM error, so we normalized the memory usage of \proposed and \ondemand to \prefetch.
}
\label{fig:peak_gpu}
\end{figure}

\subsection{Model Accuracy}
\label{sect:eval:accuracy}

In this subsection, we quantify the impact of our pre-gate function on MoE's model accuracy.
\tab{tab:switch_acc} compares the model accuracy of SwitchTransformer with and without our pre-gate function employed for various downstream tasks.
In Switch-Base with 8 experts, which is the smallest size among our studied model configurations, \proposed consistently exhibits slightly higher model accuracy across all downstream tasks.
As the model size is increased with larger number of experts, \proposed incurs a small accuracy degradation for some of the downstream tasks, but overall it continues to deliver competitive model accuracy results.
Nevertheless, this magnitude of observed variances in accuracy does not signify a substantial improvement or deterioration in the model's fundamental capabilities.
A detailed analysis on why our pre-gate function improves some of the downstream task's model accuracy is beyond the scope of this work. In general, \proposed's robust model accuracy observed across different model sizes and different downstream tasks underscores the algorithmic robustness of our proposal.

\begin{table}[t]
    \centering
    \caption{Effect of our pre-gate function on the model accuracy of Google's SwitchTransformer. R1 and R2 represent the Rouge-1 and Rouge-2 scores, respectively. For all score metrics, higher is better.}
\scriptsize
    \begin{tabular}{|c||c|c|c|c|c|c|}
    \hline
    \multirow{2}{*}{}
     & \multicolumn{2}{c|}{Xsum} & \multicolumn{2}{c|}{CB Web QA} & \multicolumn{2}{c|}{SQuAD} \\
    \cline{2-7}
     & R1 & R2  & ExactMatch & F1 & ExactMatch & F1 \\
    \hline
    Base-8 & 34.6 & 13.0 & 26.0 & 30.9 & 77.4 & 85.8 \\
    \hline
    Pre-gated & \textbf{34.7} & \textbf{13.0} & \textbf{28.2} & \textbf{32.6} & \textbf{78.2} & \textbf{86.0} \\
    \hline
    \hline
    Base-128  & \textbf{38.1} & \textbf{16.6} & \textbf{27.4} & \textbf{33.1} & 81.7 & 89.2 \\
    \hline
    Pre-gated & 38.0 & 16.5 & 25.8 & 32.2 & \textbf{82.2} & \textbf{89.4} \\
    \hline  
    \hline  
    Large-128 & \textbf{40.2} & \textbf{18.8} & \textbf{31.0} & \textbf{36.5} & \textbf{82.4} & 90.1 \\
    \hline  
    Pre-gated  & 40.1 & 18.6 & 30.5 & 36.2 & 81.9 & \textbf{90.2} \\
    \hline  
    \end{tabular}
    \label{tab:switch_acc}
\end{table}

It is important to emphasize that fine-tuning for both \proposed and conventional MoE is done using \emph{the same pre-trained model parameters with the same number of training iterations}.
The fact that \proposed produces comparable model accuracy under these conditions demonstrates the robustness of our proposal. Furthermore, it also shows that \emph{our proposal can effectively utilize pre-existing resources and training/fine-tuning recipes for deployment} (e.g., pre-trained model parameters from conventional MoE models), enhancing its applicability.

\subsection{Discussion}
\label{sect:eval:discussion}

\textbf{Pre-gating to activate experts at different blocks.} 
We have so far assumed that our pre-gate function is trained to preemptively select the experts to activate for the next subsequent MoE block. In other words, the pre-gate function's activation level ($N$) is a \emph{single} ($N$=$1$) MoE block ahead of the current MoE block. 
To explore potential optimizations in the MoE architecture using our pre-gate function, we evaluate the model accuracy of MoE when the pre-gate function is trained to select the experts to activate for the 2nd/3rd subsequent MoE block ahead ($N$=$2$/$3$), the result of which is shown in  \fig{fig:pre_gate_level}.

\begin{figure}[t!] \centering
\includegraphics[width=0.43\textwidth]{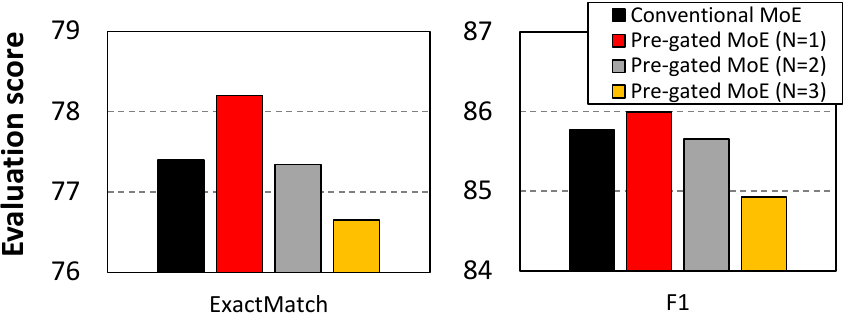}
\caption{Effect on model accuracy when changing the activation level of our pre-gate function, from a single MoE block ahead ($N$=$1$, our default configuration) to 2nd/3rd MoE block ahead ($N$=$2$/$3$). Evaluation is conducted over Switch-Base with $8$ experts for the closed-book question answering task trained with the SQuAD dataset (ExactMatch (left) and F1 (right) scores are the evaluation metrics for the given task).} 
\label{fig:pre_gate_level}
\end{figure}

As depicted, our default \mbox{\proposed} configuration (pre-gating with activation level-1, i.e., $N$=1) in the Switch-Base model with 8 experts consistently shows the highest model accuracy than the rest of the design points including conventional MoE structure (i.e., selecting experts to activate for the \emph{current} MoE block, $N$=$0$) as well as pre-gate functions trained to select 2nd/3rd subsequent MoE block ahead ($N$=$2$/$3$).
Note that the model accuracy gradually decreases as the pre-gate function's activation level increases (from $N$=$1$ to $3$). We conjecture that the further away the preemptively selected MoE block is from the current pre-gate function, the less likely the current pre-gate function's input activations will contain useful information to accurately select what experts are most suitable to activate. A detailed evaluation of such is beyond the scope of this work and we leave it as future work.

\begin{figure}[t!] \centering
\includegraphics[width=0.49\textwidth]{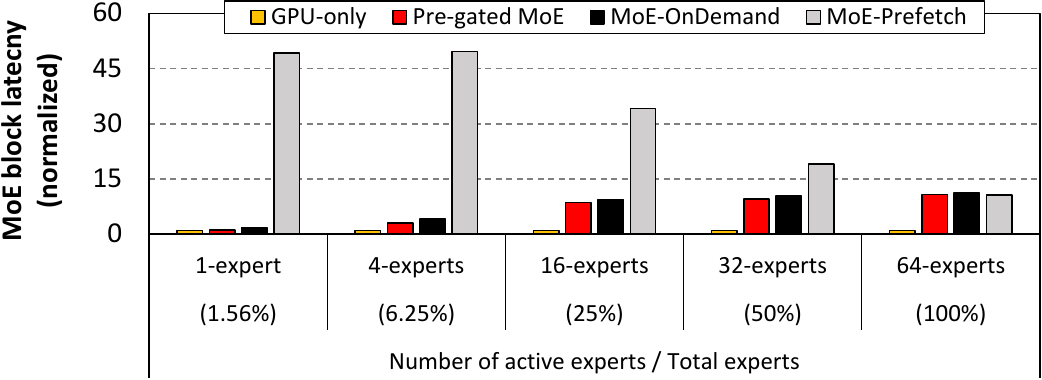}
\caption{Effect of the number of activated experts on MoE block latency (normalized to \gpuonly). Evaluation is conducted using Switch-Base with $64$ experts.} 
\label{fig:ubench}
\end{figure}

{\bf Number of experts activated.} The power of MoE comes from its \emph{sparse} activation of experts (designed to mimic the behavior of the human brain, i.e., specialize regions of the brain tuned for specific tasks), which allows the model architecture to scale its model capacity without proportionally increasing its computational demand. For example, the default model configuration of Google's SwitchTransformer activates just a single expert (top-1 activation) in a single batch inference, so a SwitchBase model with $64$ experts will activate only $1.56\%$ of its experts. For the completeness of our study, we show in \fig{fig:ubench} the performance of \proposed when we manually increase the number of activated experts in Switch-Base with 64 experts from $1$ expert ($1.56\%$ expert activation) to $64$ experts ($100\%$ expert activation). 

There are two key observations that can be made from this experiment. First, all CPU offloading based solutions (\proposed, \ondemand, and \prefetch) experience a  higher performance degradation vs. \gpuonly as the number of activated experts is increased.
This is expected because the behavior of MoE becomes similar to a dense LLM model when a larger number of experts are activated (i.e., all model parameters are utilized with $100\%$ activation), rendering CPU offloading solutions less effective.
Second, the performance gap between \prefetch and \proposed gradually reduces as the number of activated experts increases. Because \prefetch migrates the entire expert parameters for every MoE block, a larger number of activated experts reduces the needlessly \emph{overfetched} expert parameters, closing its performance gap against \proposed. Nonetheless, for MoE models with sparse expert activations (the most common way of developing an MoE model architecture), \proposed demonstrates its robustness and consistently provides superior performance than other \gpucpu systems.

\begin{figure}[t!] \centering

\includegraphics[width=0.49\textwidth]{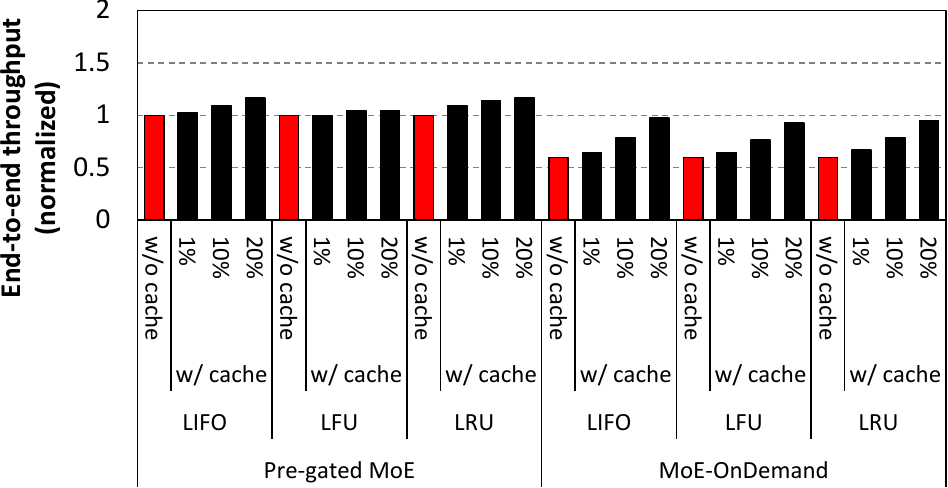}

\caption{
End-to-end throughput of \proposed and \ondemand when evaluated with the Switch-Large (128 experts) model (normalized to \proposed without caching). When caching is enabled, we change the fraction of experts that are cached inside the GPU memory (from $1\%$ to $20\%$) and compare its effectiveness.
For the completeness of our study, we evaluate not only the LIFO policy suggested by \cite{meta_moe} but also a least frequently used (LFU) replacement policy~\cite{semoe} and a least-recently used (LRU) replacement policy.
}
\vspace{-0.5em}
\label{fig:cache}
\end{figure}

\textbf{Caching experts on \proposed.}
Prior work by Huang et al.~\cite{meta_moe} characterized MoE models for machine translation and language modeling, uncovering the existence of a few hot active experts during inference. Based on such observation, \cite{meta_moe} explores \emph{expert buffering} for MoE inference which caches hot, active experts in GPU memory using a last in first out (LIFO) cache replacement policy, while buffering the rest in CPU memory.
To evaluate the effectiveness of expert caching~\cite{meta_moe,semoe} on our \proposed as well as other CPU offloading based MoE designs, we implement a caching system on top of both \proposed and \ondemand and evaluate its performance. 

As shown in \fig{fig:cache}, caching experts generally provides performance benefits to both \proposed and \ondemand, regardless of the types of the cache replacement policy employed. However, the effectiveness of caching is more pronounced with \ondemand as the performance overhead incurred with expert migration is more severe under this design point, unlike \proposed which is already capable of hiding most of the expert migration latency by overlapping it with expert execution.

\textbf{\proposed with SSD offloading.}
Prior work~\cite{semoe} evaluates the efficacy of offloading MoE parameters to SSDs as means to deploy even larger LLMs. To evaluate the effectiveness of \proposed on top of such design point, we implement \proposed and all baseline systems on top of an SSD offloading based MoE serving system, the result of which is summarized in \fig{fig:ssd_offloading}. As depicted, the performance benefit of \proposed against other baseline systems is decreased compared to a CPU ``DRAM'' offloaded MoE system. This is because, when the MoE parameters are offloaded to an SSD, the expert migration latency between SSD$\rightarrow$GPU becomes much longer compared to migrating it from CPU DRAM (due to the much lower slower data transfer bandwidth between SSD vs. CPU DRAM). Consequently, the expert migration latency becomes such a huge end-to-end performance bottleneck that it completely overwhelms the overall system, rendering the effectiveness of any CPU offloading based approaches to become smaller. Nonetheless, \proposed consistently delivers higher performance than all other baseline systems demonstrating its robustness. 

\begin{figure}[t!] \centering
\includegraphics[width=0.49\textwidth]{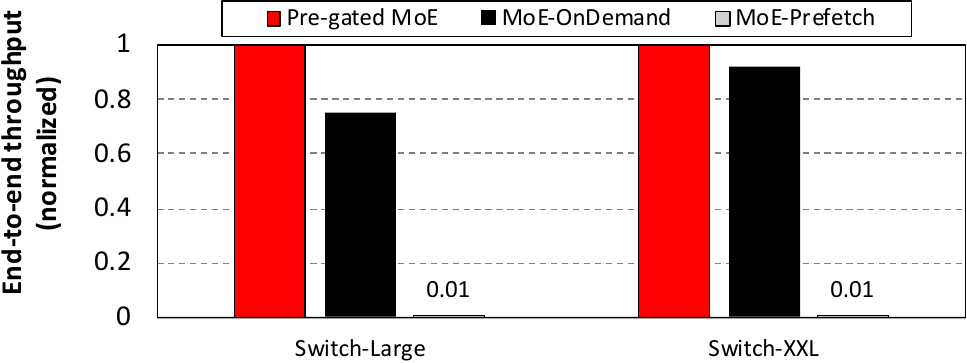}
\caption{End-to-end inference throughput of SSD offloading. In this experiment, we additionally evaluate a larger MoE model named Switch-XXL, a SwitchTransformer based model architecture that has the identical 
configuration as Switch-Large but increases both the feature vector dimension size and the number of heads by $4\times$, amounting to 395 billion parameters ($16\times$ more than Switch-Large) and 217 GB in model size after quantization is applied.
\gpuonly suffers from an OOM error, so performance is normalized to \proposed. 
} 
\vspace{-0.5em}
\label{fig:ssd_offloading}
\end{figure}

\section{Related Works} 
\label{sect:related}

There exists a large number of prior work exploring ML inference systems for  MoE-based LLMs~\cite{tamoe, fastmoe, fastermoe, tutel, semoe, dsmoe, linamoe, smartmoe}. In this section, we summarize prior work by categorizing them into three different categories: (1) systems for MoE training, 2) systems for MoE inference, and 3) MoE model architectures for efficient MoE deployment.

\textbf{Systems for the MoE training.}
Prior work on FastMoE~\cite{fastmoe} and FasterMoE~\cite{fastermoe} propose system-level optimizations for multi-GPU solutions, specifically tackling the load-imbalance issue in MoE training.
Tutel~\cite{tutel} presents dynamic multi-GPU parallelism and pipelining optimization for distributed MoE training systems.
SmartMoE~\cite{smartmoe} 
explores efficient search strategies for parallelizing MoE training. 
TA-MoE~\cite{tamoe} and Li et al.~\cite{linamoe} propose optimizations for MoE training's all-to-all communication and expert routing.
Unlike \proposed which focuses on inference, all of these prior works concentrate on MoE training over multi-GPU systems, assuming all model parameters are partitioned across the GPUs allowing each model partition to be stored in GPU memory.

\textbf{Systems for the MoE inference.}
DeepSpeed-MoE~\cite{dsmoe} and Li et al.~\cite{linamoe} propose efficient communication optimizations as well as compute kernel optimizations for multi-GPU based MoE inference systems. 
DeepSpeed-inference~\cite{dsinf} proposes to offload memory hungry tensors (e.g., activations, parameters) to the CPU memory and NVMe SSD following ZeRO-offload~\cite{zero_offload} and ZeRO-infinity~\cite{zero_infinity}. DeepSpeed-inference, however, did not evaluate their parameter offloading feature to sparse MoE architectures targeting the memory capacity limited expert parameters. HuggingFace Accelerate~\cite{huggingface_accelerate} and SE-MoE~\cite{semoe} respectively implement the \ondemand and \prefetch systems we evaluate in this paper, a CPU offloading based MoE inference system.

\textbf{Efficient MoE model architectures.}
DeepSpeed-inference~\cite{dsinf} proposed PR-MoE and Mixture-of-Student (MoS) architectures, which help significantly compress down the model size of MoE.
However, these models require significant modifications to the model architecture based on knowledge distillation and often result in model accuracy degradation. Furthermore, these models are designed for GPU-only configurations, unlike the CPU offloading based \proposed.
SE-MoE~\cite{semoe} also proposed a compact MoE model architecture based on distillation, compression, and pruning, but it suffers from non-negligible degradation in model accuracy.
Our \proposed, on the other hand, only requires modest changes to the MoE model architecture without compromising model accuracy.

\section{Conclusion}
\label{sect:conclusion}

This paper presents \proposed, our algorithm-system co-design
for scalable and high-performance MoE inference.
\proposed effectively addresses the two main challenges of MoE (its large memory footprint and dynamic nature of sparse expert activation) via our novel pre-gate function, 
which alleviates the dynamic nature of sparse expert activation,
allowing our proposed system to address the large memory
footprint of MoEs while also achieving high performance. 
Compared to state-of-the-art MoE inference systems, \proposed improves inference throughput while significantly reducing the GPU memory consumption. Importantly, \proposed offers comparable model accuracy across various natural language processing tasks, facilitating its adoption in a wide range of real-world applications.

\section*{Acknowledgment}
This research was supported by the MSIT (Ministry of Science, ICT), Korea, under the High-Potential Individuals Global Training Program (RS-2022-00155958) supervised by the IITP (Institute for Information \& Communications Technology Planning \& Evaluation)
and this project is (partially) supported by Microsoft Research Asia.

\bibliographystyle{IEEEtranS}
\bibliography{refs}

\newpage

\appendix
\section{Artifact Appendix}

\subsection{Abstract}

This artifact evaluation repository contains the base implementation of Pre-gated MoE. The artifact is designed to demonstrate the performance of our Pre-gated MoE compared to the three baselines mentioned in this paper. As detailed in \sect{sect:methodology}, we built our Pre-gated MoE on top of Google’s SwitchTransformer~\cite{switchtransformer}, utilizing NVIDIA’s FasterTransformer~\cite{fastertransformer}.

\subsection{Artifact check-list (meta-information)}

{\small
\begin{itemize}
  \item {\bf Algorithm: } Pre-gated MoE algorithm
  \item {\bf Program: } C++, Python
  \item {\bf Model: } Google's SwitchTransformer
  \item {\bf Hardware: } At least one GPU with 40GB of memory and a CPU with 128GB of memory.
  \item {\bf Output: }  Key results of our paper including average MoE block latency, inference throughput, and peak GPU memory consumption.
  \item {\bf How much disk space required (approximately)?: } Over 100GB for storing the model parameters.
  \item {\bf How much time is needed to prepare workflow (approximately)?: } 6 hours
  \item {\bf How much time is needed to complete experiments (approximately)?: } 30 minutes
  \item {\bf Publicly available?: } Yes
  \item {\bf Archived: } \url{https://doi.org/10.5281/zenodo.10976343}
\end{itemize}
}

\subsection{Description}

\subsubsection{How to access}
The artifact is available in archival repositories on Zenodo and GitHub.
\begin{itemize}
    \item Zenodo: \url{https://doi.org/10.5281/zenodo.10976343}
    \item GitHub: \url{https://github.com/ranggihwang/Pregated_MoE}
\end{itemize}

\subsubsection{Hardware dependencies}
To reproduce the results presented in the paper, the following hardware is required:
\begin{itemize}
    \item A CPU with at least 128GB of memory.
    \item A GPU with at least 40GB of memory. (We recommend using recent GPUs, such as the NVIDIA A100 with 80GB HBM, for optimal performance.)
    \item Additionally, ensure there is more than 100GB of disk storage available for model parameters.
\end{itemize}

\subsubsection{Software dependencies}

We advise using the Docker image following the description in our repository to circumvent most software-related issues. The repository includes a script that automates the installation of all necessary software dependencies for compiling and running the artifact. Further details are provided in the repository documentation.

\subsubsection{Data sets}

To use the artifact, it is necessary to download the model weights for the SwitchTransformer from HuggingFace. Our repository provides detailed instructions and scripts for downloading these model weights.

\subsection{Installation}
\begin{enumerate}
    \item Create a directory for model preparation.

    \item Launch a Docker container using the following command, replacing \texttt{ \$$\{$DATA\_PATH$\}$} with the path to your model preparation directory:

    \begin{lstlisting}
    docker run -ti --gpus all --shm-size 5g --name pregated -v ${DATA_PATH}:/data nvcr.io/nvidia/pytorch:22.09-py3 bash
    \end{lstlisting}

    \item Clone the repository and initiate the build process. The \texttt{-DSM} parameter should match your GPU’s compute capability. Refer to the documentation to select the appropriate value for your setup.
   
    \begin{lstlisting}
    # build on A100
    mkdir -p FasterTransformer/build
    cd FasterTransformer/build
    cmake -DSM=80 -DCMAKE_BUILD_TYPE=Release -DBUILD_PYT=ON -DBUILD_MULTI_GPU=ON ..
    make -j
    \end{lstlisting}

    \item  Install the required Python dependencies:

    \begin{lstlisting}
    pip install -r ../examples/pytorch/t5/requirement.txt
    \end{lstlisting}

\end{enumerate}

\subsection{Experiment workflow}
\begin{enumerate}
    \item Prepare the models.
    \begin{lstlisting}
    mkdir /data/ft
    cd /workspace/FasterTransformer/
    ./scripts/convert.sh
    \end{lstlisting}

    \item Begin the evaluation using the provided script:
    
    \begin{lstlisting}
    cd /workspace/FasterTransformer/
    # logs will be output here
    mkdir logs/
    python scripts/eval_all.py
    \end{lstlisting}

\end{enumerate}

\subsection{Evaluation and expected results}

The script will generate output files in CSV format including \texttt{block\_lats.csv}, \texttt{throughputs.csv} and \texttt{peak\_mems.csv}, which contain data on MoE block latencies, inference throughputs, and peak memory usage, respectively. The results are shown in \fig{fig:moe_block_latency_model}, \fig{fig:throughput_all}, and \fig{fig:peak_gpu}.

\subsection{Experiment customization}
To customize the experiment, you can modify the evaluation configuration in  \texttt{scripts/eval\_all.py}

\subsection{Notes}

\subsection{Methodology}

Submission, reviewing and badging methodology:

\begin{itemize}
  \item \href{https://www.acm.org/publications/policies/artifact-review-and-badging-current}{https://www.acm.org/publications/policies/artifact-review-and-badging-current}
  \item \href{http://cTuning.org/ae/submission-20201122.html}{http://cTuning.org/ae/submission-20201122.html}
  \item \href{http://cTuning.org/ae/reviewing-20201122.html}{http://cTuning.org/ae/reviewing-20201122.html}

\end{itemize}

\end{document}